\documentclass{svproc}
\usepackage{url}
\usepackage{graphicx}
\usepackage{amsmath}
\usepackage{booktabs}
\usepackage{array}

\newcolumntype{P}[1]{>{\centering\arraybackslash}p{#1}}

\begin{document}
\mainmatter              
\title{FasteNet: A Fast Railway Fastener Detector}
\titlerunning{FasteNet Railway Fastener Detection}
\author{Jun Jet Tai\inst{1} \and Mauro S. Innocente\inst{1} \and Owais Mehmood\inst{2}}
\authorrunning{Jun Jet Tai et al.} 

\institute{Coventry University, Priory St, Coventry CV1 5FB, UK\\
\email{taij@uni.coventry.ac.uk},\\
\email{Mauro.S.Innocente@coventry.ac.uk},
\texttt{https://availab.org/}
\and
Omnicom Balfour Beatty, Clifton Park Ave, Rawcliffe, York YO30 5PB, UK\\
\email{owais.mehmood@balfourbeatty.com}\\
}

\maketitle              

\begin{abstract}
In this work, a novel high-speed railway fastener detector is introduced. This fully convolutional network, dubbed FasteNet, foregoes the notion of bounding boxes and performs detection directly on a predicted saliency map. Fastenet uses transposed convolutions and skip connections, the effective receptive field of the network is 1.5$\times$ larger than the average size of a fastener, enabling the network to make predictions with high confidence, without sacrificing output resolution. In addition, due to the saliency map approach, the network is able to vote for the presence of a fastener up to 30 times per fastener, boosting prediction accuracy. Fastenet is capable of running at 110 FPS on an Nvidia GTX 1080, while taking in inputs of 1600$\times$512 with an average of 14 fasteners per image. Our source is open here: https://github.com/jjshoots/DL\_FasteNet.git

\end{abstract}

\section{Introduction}
Traditionally, the inspection of railway fasteners has been done by visual inspection \cite{marquis2014vehicle}. However, modern inspection methods are taking over this task. Examples include the Viola-Jones algorithm \cite{xia2010broken} and symmetry-based pyramid histogram of oriented gradients (PHOG) algorithm \cite{liu2015integrating}. Gibert et. al. used support vector machines (SVMs) to perform fastener and defect detection \cite{gibert2015robust}. The work was also extended to fully convolutional neural networks \cite{gibert2016deep}. Comparisons show that deep learning methods work well even with limited training data while being easier to design. More crucially, deep learning models tend to have lower inference time and good generalizability. Recent work by Song et. al. \cite{song2019high} utilized a ResNet18 \cite{he2016deep} backbone network on top of Faster-RCNN \cite{ren2015faster}. Their implementation achieved a recall and precision of more than 99\% but was capped at 35 FPS on an Nvidia Titan X for 210 fasteners detected per second. Wang et. al. proposed using the popular YOLOv2 and YOLOv3 architectures to perform fastener component detection \cite{wang2020real}. Their implementation on an Nvidia GTX 1080Ti worked on close up images of railway fasteners, and performed with a mean average precision (mAP) of 93\% at 35 FPS. These works show that deep convolutional neural networks (DCNNs) can be applied to railway fastener detection, like other computer vision tasks before it. However, the task of fastener, defect and component detection is unique. Generally, the scale and rotation of the fasteners do not change, there is also no overlap between fasteners. This is true even in the case of sleeper detection. This is in contrast to general object detection tasks that may be subject to scale variance, occlusion, overlap, and other visual interruptions. For this reason, many existing network architectures are inelegant and can be improved. This work introduces several novel ideas:

\begin{itemize}
    \item Bounding boxes prevent the network from making more than one prediction per fastener without output suppression. Even so, predicting more than 10 bounding boxes at every output spatial location is excessive. FasteNet instead predicts a saliency map. The bounding boxes are then derived from this output using a first principles method, boosting accuracy by allowing the network to predict up to 30 times per fastener without requiring more than a one channel output feature map.
    \item There is no need to train the network to work on a large range of fastener scales. Due to this, FasteNet has an effective receptive field (ERF) only around 50\% larger than necessary.
    \item FasteNet has different regimes during training and inference. During inference, outputs are thresholded and ceiled to increase prediction accuracy by eliminating low confidence predictions. During training, this step is removed to allow the network to fully train for every output location.
\end{itemize}

The rest of this paper is arranged as follows; in Section \ref{methodology}, our proposed system architecture is presented. In Section \ref{performance_analysis}, the novelties that improve system performance are introduced. Finally, the paper is concluded in Section \ref{conclusion}.

\section{Methodology} \label{methodology}
\subsection{Problem Description}
\begin{figure} [ht]
    \centering
    (a) \includegraphics[width=0.45\textwidth]{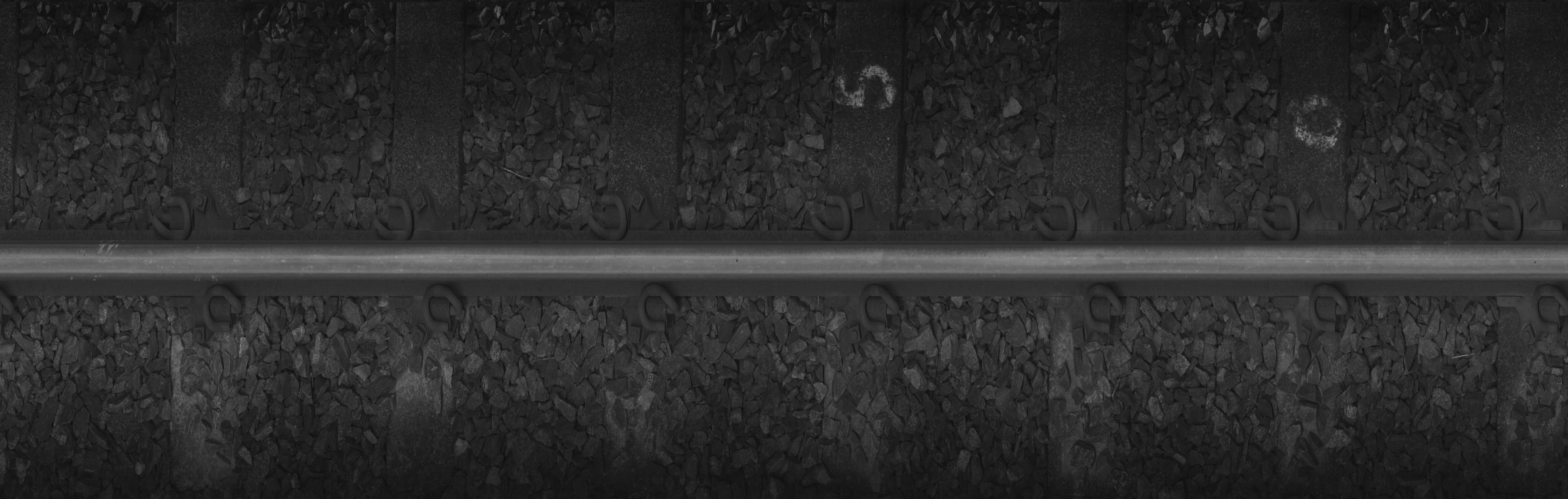}
    (b) \includegraphics[width=0.45\textwidth]{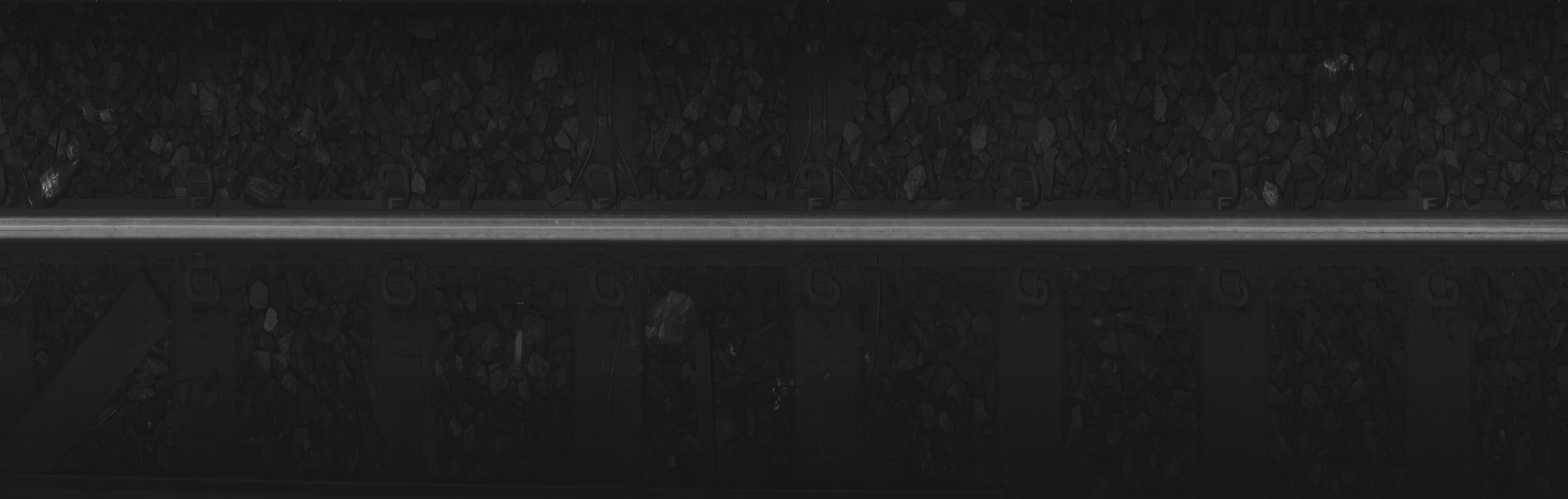}
    \caption{Example of railway images in the dataset where (a) all fasteners are visible, (b) the image is very badly illuminated.}
    \label{fig:fastener_images}
\end{figure}

FasteNet's purpose is to count fasteners in an image; not novel, but serves as a proof of concept for the ideas introduced here. A set of 997 images are generated by a New Measurement Train, some examples are shown in Fig. \ref{fig:fastener_images}. The fasteners in the images include fasteners which are occluded, damaged or missing. Furthermore, the fasteners are exposed under various lighting conditions due to time of day; no artificial illumination is used. The images also contain fasteners of various types - PR clip, E clip, Fast clip, C clip, and J clip. The images are solely black and white, making detection harder by removing colour information.

The problem is treated as a localization one. As long as the network prediction area overlaps with a ground truth bounding box, the prediction is considered a true positive (TP), otherwise it is perceived as a false positive (FP).
The network must output the number of fasteners it sees in the image. This is to enforce the idea that the algorithm must clearly distinguish one fastener from another fastener, as opposed to simply outputting a saliency map. This variation of object detection strays away from the general bounding box detection one \cite{fu2017dssd} \cite{redmon2018yolov3} \cite{ren2015faster}. As long as the the prediction area lies over a real fastener is sufficient.

\subsection{FasteNet Design}

\begin{figure} [ht]
    \centering
    \includegraphics[width=\textwidth]{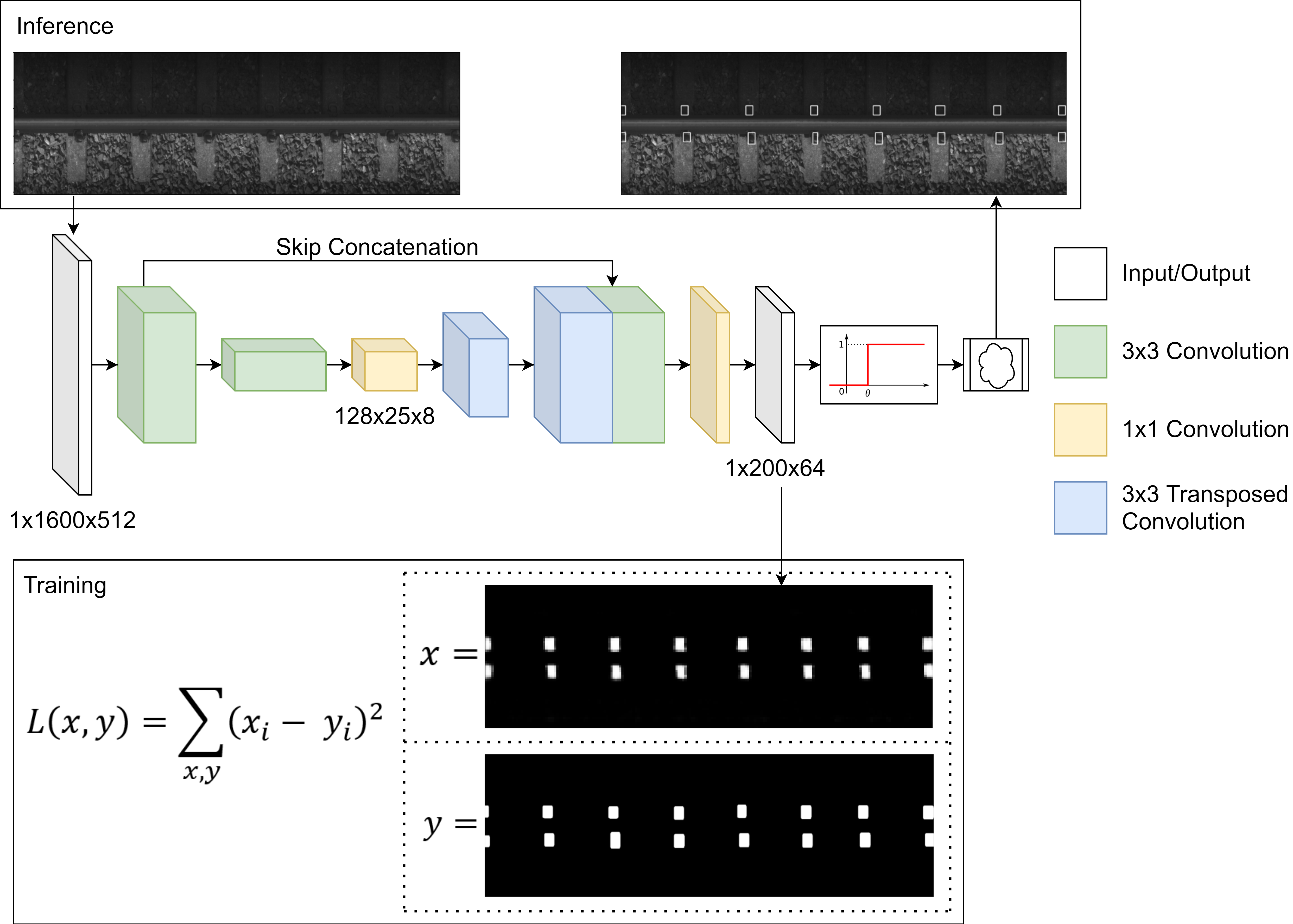}
    \caption{Top level view of FasteNet. During training, the network trains against segmentation masks. During inference, the network outputs are thresholded before a contour finding algorithm is applied to find individual fasteners.}
    \label{fig:FasteNet}
\end{figure}

Fig. \ref{fig:FasteNet} provides a top level overview of FasteNet's architecture, as well as training and inference regimes. The input to the network is a black-and-white image defined here as $w_i \in \rm I\!R^{1 \times m \times n}$ where $m \leftrightarrow 64 | m$  and $n \leftrightarrow 64 | n$. It outputs a saliency map of $x_i \in \rm I\!R^{1 \times \frac{m}{8} \times \frac{n}{8}}$. The backbone network of FasteNet has just-enough layers to provide it with a roughly 64$\times$64 ERF. This is a slightly hand-wavy method of calculating the ERF; in reality, the ERF more closely represents a Gaussian distribution on the true receptive field \cite{araujo2019computing}. This ERF is roughly 1.5$\times$ the size of a fastener in the image. However, using only convolutional and pooling layers to achieve this ERF would translate to a very poor output resolution of 1/64th the input resolution. Many implementations of object detection work well with low output resolutions by using multiple bounding boxes per spatial element, and then suppressing repeat predictions \cite{redmon2017yolo9000} \cite{fu2017dssd}. This approach relies on very deep networks to provide the the output layers with enough semantic value, and then uses bounding boxes to handle localization precision loss.

Our approach takes a different look at the problem. FasteNet uses transposed convolutions to increase the output spatial resolution to 200$\times$64 (1/8th the input size), and then predicts an objectness score at each output location. To reduce information loss and alleviate the vanishing gradient problem \cite{hochreiter1998vanishing}, feature maps from an earlier layer are concatenated to layers before the output layer. 1$\times$1 convolutions are then used to produce the output saliency map. This allows up to 30 predictions per fastener. Other tidbits that boost FasteNet's performance include batch normalization to combat internal covariate shift during training and to provide regularization. Leaky ReLU units with a negative gradient of -0.1 are used to prevent dead neurons. The overall network architecture is shown in Table \ref{tab:architecture}, note that the spatial size depends on the input size; any input size divisible by 64 can be computed by FasteNet - our implementation uses an input size of 1600$\times$512 \textit{during inference}.


During training, FasteNet trains against semantic segmentation masks. An example is shown in Fig. \ref{fig:FasteNet}. During inference, FasteNet thresholds all output pixels, $x_i \leftarrow \delta_{[x > \theta]}$, where $\delta$ is the Kronecker delta, and then uses a boundary walking algorithm by Suzuki et. al. \cite{suzuki1985topological} to identify each cluster at the output. The cluster bounds are used to derive bounding boxes. Semantically, given a chain of points $c_i \in \rm I\!R^{n \times 2}$ denoting the contour for a cluster, the top left $\mathbf{b}_\text{1}$ and bottom right $\mathbf{b}_\text{2}$ corners of the bounding boxes are defined as $\mathbf{b}_\text{1} = \{\min_{i \in n} \mathbf{c}_{i_x}, \max_{j \in n} \mathbf{c}_{j_y}\}$ and $\mathbf{b}_\text{2} = \{\max_{i \in n} \mathbf{c}_{i_x}, \min_{j \in n} \mathbf{c}_{j_y}\}$. Other algorithms for drawing bounding boxes can also be used. In an extreme case, bounding boxes can be foregone exclusively, and the shape of the contour itself can be used to identify instances of fasteners.

\subsection{Training and Evaluation} \label{training}
The network is trained with 1,000 256$\times$256 random crops per image for 700 out of 997 available images. The mean squared error (MSE) objective is used, $L(\mathbf{x}, \mathbf{y}) = \sum_{\mathbf{x}, \mathbf{y}}(x_i - y_i)^2$, where $\mathbf{x}$ and $\mathbf{y}$ represent the network prediction and ground truth mask respectively. The Adam optimizer \cite{kingma2014adam} with a learning rate of 1e-6 and weight decay of 1e-2 is set as the optimizer. The network converges after the first 20 epochs with a precision of 96\% and a recall of 89\% when validated on the remaining 297 images. Hard negative mining was then used to train the network on difficult training images (more than 1 FP or FN). The network was then explicitly trained on these examples for 2 epochs, before continuing on the whole 700 images for another 2 epochs. This was repeated 3 times. In the end, a precision of 99\% and a recall of 90\% was accomplished on the validation dataset.


\begin{table}[ht]
\scriptsize
\centering
\caption{Network architecture of FasteNet. The outputs from layer 3 are concatenated channel wise to outputs of layer 10 to form the input to layer 11.}
\label{tab:architecture}
\resizebox{\textwidth}{!}{%
\begin{tabular}{@{}p{1cm}|p{1.5cm}|p{2cm}|p{2cm}|p{1cm}|p{1cm}|p{1cm}|l@{}}
\hline
Layer \# & Operation              & Input Size        & Output Size       & Kernel Size & Pad Size                      & Pool Size   & Activation \\ \hline
1        & Convolution            & 1@ 1600$\times$512 & 32@ 800$\times$256 & 3$\times$3  & 1$\times$1  & 2$\times$2 & Leaky ReLU \\
2        & Convolution            & 32@ 800$\times$256 & 32@ 400$\times$128 & 3$\times$3  & 1$\times$1  & 2$\times$2 & Leaky ReLU \\
3        & Convolution            & 32@ 400$\times$128 & 64@ 200$\times$64  & 3$\times$3  & 1$\times$1  & 2$\times$2 & Leaky ReLU \\ \hline
4        & Convolution            & 64@ 200$\times$64  & 64@ 100$\times$32  & 3$\times$3  & 1$\times$1  & 2$\times$2 & Leaky ReLU \\
5        & Convolution            & 64@ 100$\times$32  & 128@ 50$\times$16  & 3$\times$3  & 1$\times$1  & 2$\times$2 & Leaky ReLU \\
6        & Convolution            & 128@ 50$\times$16  & 128@ 25$\times$8   & 3$\times$3  & 1$\times$1  & 2$\times$2 & Leaky ReLU \\
7        & Convolution            & 128@ 25$\times$8   & 64@ 25$\times$8    & 1$\times$1  & -           & -          & Leaky ReLU \\
8        & Transposed convolution & 64@ 25$\times$8    & 64@ 50$\times$16   & 4$\times$4  & 1$\times$1  & -          & Leaky ReLU \\
9        & Transposed convolution & 64@ 50$\times$16   & 64@ 100$\times$32  & 4$\times$4  & 1$\times$1  & -          & Leaky ReLU \\
10       & Transposed convolution & 64@ 100$\times$32  & 64@ 200$\times$64  & 4$\times$4  & 1$\times$1  & -          & Leaky ReLU \\ \hline
11       & Convolution            & 128@ 200$\times$64 & 1@ 200$\times$64   & 1$\times$1  & -           & -          & Sigmoid    \\ \hline
\end{tabular}%
}
\end{table}


\section{Performance Analysis} \label{performance_analysis}
The model is validated on the remaining 297 images from the dataset. Evaluation with metrics done by other authors \cite{gibert2016deep} \cite{gibert2015robust} \cite{liu2019learning} \cite{wei2019railway} is not possible as the datasets are not public and the detection classes are different. Instead, FasteNet is compared to two other internal models. This is to validate our hypothesis that a network designed in this manner has better performance than a brute force DCNN implementation. These two networks are denoted here as VanillaNet and LargeNet, shown in Table \ref{tab:vanillanet} and Table \ref{tab:largenet} with a comparison in Table \ref{tab:network_comparisons}.

VanillaNet does not utilize transposed convolutions and consists of 11 layers. LargeNet uses transposed convolutions with a larger ERF of roughly 128$\times$128. Note that because of the output scaling ratio of this network, the inputs are cropped to 1536$\times$512. These two additional networks are trained on the same training regime that was used on FasteNet in Section \ref{training}. Special considerations are made to ensure that each network has roughly the same number of parameters to reduce variance in representational capacity.

\begin{figure} [!ht]
    \centering
    \includegraphics[width=0.8\textwidth]{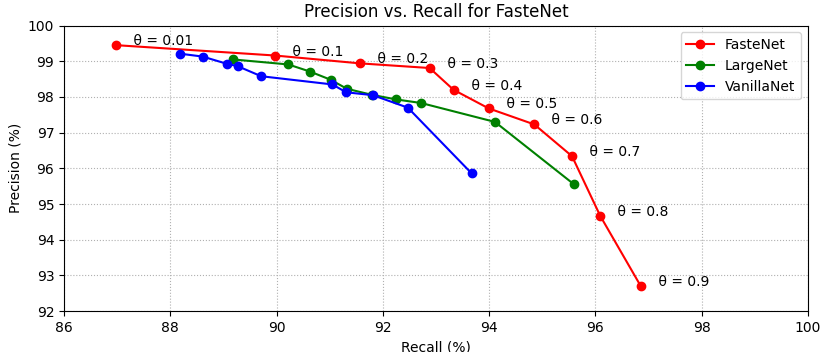}
    \caption{Precision Recall Curve of Fastenet, VanillaNet, and LargeNet when performed on the unseen 297 images of the dataset with different threshold values.}
    \label{fig:pr_curve}
\end{figure}

\begin{figure} [!ht]
    \centering
    (a) \includegraphics[width=0.45\textwidth]{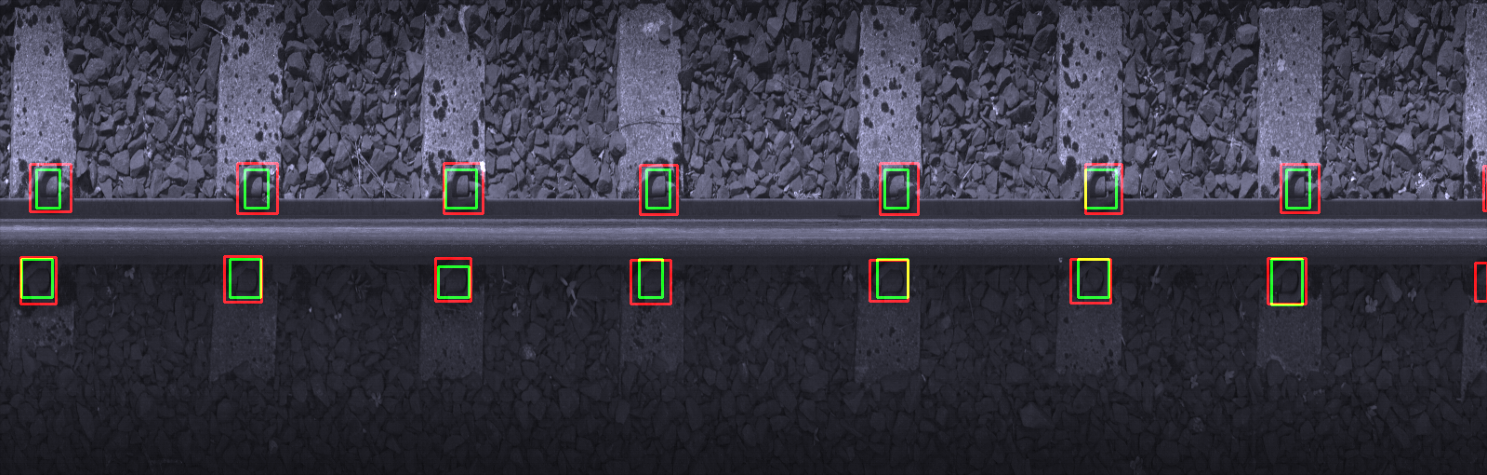}
    (b) \includegraphics[width=0.45\textwidth]{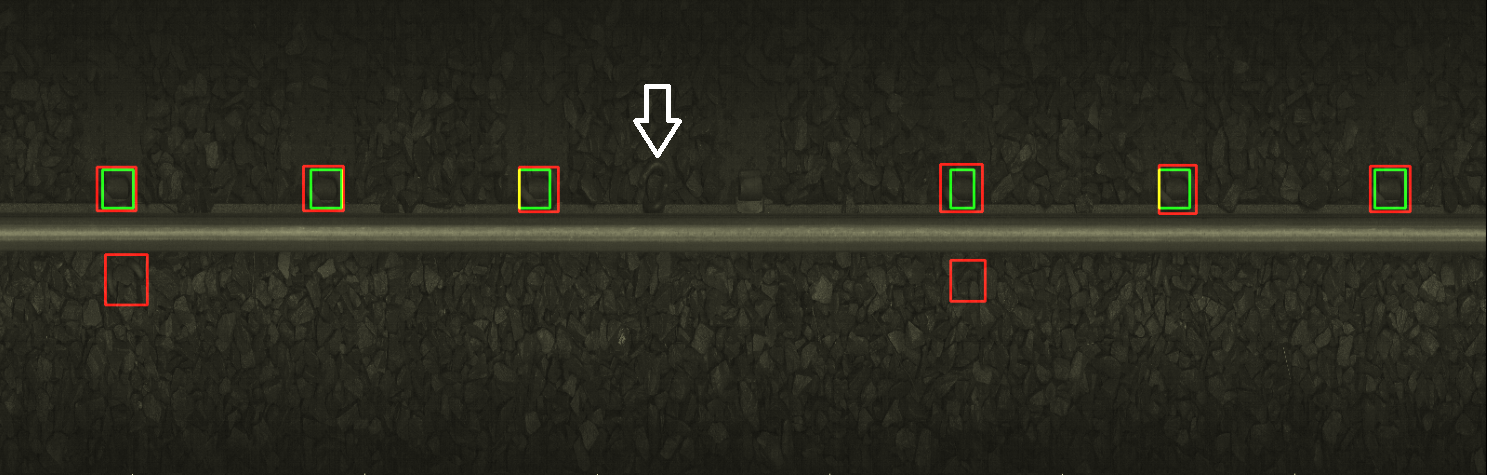}
    \caption{Green boxes are FasteNet predictions, red boxes are ground truth annotations. (a) 'False' Positives, the network predicts the presence of the fasteners even if small and not annotated in the ground truth. (b) False Negatives resulting from fastener occlusion and poor lighting conditions. Even so, the network hedges against detecting defect fasteners, identified with the white arrow here.}
    \label{fig:network_outputs}
\end{figure}

By varying the threshold $\theta$, the rate of predictions can be varied to form the precision recall curves for each network shown here in Fig. \ref{fig:pr_curve}. The performance of FasteNet is on top. LargeNet and VanillaNet implementations may perform as well or better than FasteNet when given more training time due to deeper network architectures and hence theoretically more representational capacity. However, this sub-percentage performance increase is compensated by the loss in some computational speed, which is especially the case in VanillaNet. At the end of training, FasteNet is able to output bounding boxes around each fastener as shown in Fig. \ref{fig:network_outputs}. 

\begin{table}[ht]
\scriptsize
\centering
\caption{Summary of FasteNet in comparison with two other networks built for an evaluation of FasteNet's design.}
\label{tab:network_comparisons}
\begin{tabular}{@{}lp{2.5cm}ll@{}}
\toprule
Network    & Parameter Count & FLOPS  & FPS \\ \midrule
FasteNet   & 494k            & 4.00G  & 110 \\
VanillaNet & 500k            & 25.93G & 25  \\
LargeNet   & 514k            & 6.38G  & 80  \\ \bottomrule
\end{tabular}
\end{table}

\begin{table}[ht!]
\scriptsize
\centering
\caption{Network architecture of VanillaNet.}
\label{tab:vanillanet}
\begin{tabular}{@{}p{1cm}|p{1.5cm}|p{2cm}|p{2cm}|p{1cm}|p{1cm}|p{1cm}|l@{}}
\hline
Layer \# & Operation   & Input Size         & Output Size        & Kernel Size & Pad Size   & Pool Size  & Activation \\ \hline
1        & Convolution & 1@1600$\times$512  & 32@800$\times$256  & 3$\times$3  & 1$\times$1 & 2$\times$2 & Leaky ReLU \\
2        & Convolution & 32@800$\times$256  & 64@800$\times$256  & 3$\times$3  & 1$\times$1 & -          & Leaky ReLU \\
3        & Convolution & 64@800$\times$256  & 32@800$\times$256  & 3$\times$3  & 1$\times$1 & -          & Leaky ReLU \\
4        & Convolution & 32@800$\times$256  & 64@400$\times$128  & 3$\times$3  & 1$\times$1 & 2$\times$2 & Leaky ReLU \\
5        & Convolution & 64@400$\times$128  & 128@400$\times$128 & 3$\times$3  & 1$\times$1 & -          & Leaky ReLU \\
6        & Convolution & 128@400$\times$128 & 64@400$\times$128  & 3$\times$3  & 1$\times$1 & -          & Leaky ReLU \\
7        & Convolution & 64@400$\times$128  & 128@200$\times$64  & 3$\times$3  & 1$\times$1 & 2$\times$2 & Leaky ReLU \\
8        & Convolution & 128@200$\times$64  & 64@200$\times$64   & 3$\times$3  & 1$\times$1 & -          & Leaky ReLU \\
9        & Convolution & 64@200$\times$64   & 128@200$\times$64  & 3$\times$3  & 1$\times$1 & -          & Leaky ReLU \\
10       & Convolution & 128@200$\times$64  & 64@200$\times$64   & 3$\times$3  & 1$\times$1 & -          & Leaky ReLU \\
11       & Convolution & 64@200$\times$64   & 1@200$\times$64    & 1$\times$1  & 1$\times$1 & -          & Sigmoid    \\ \hline
\end{tabular}
\end{table}

\begin{table} [ht!]
\scriptsize
\centering
\caption{Network architecture of LargeNet. The outputs from layer 3 are concatenated channel wise to outputs of layer 12 to form the input to layer 13.}
\label{tab:largenet}
\begin{tabular}{@{}p{1cm}|p{1.5cm}|p{2cm}|p{2cm}|p{1cm}|p{1cm}|p{1cm}|l@{}}
\hline
Layer \# & Operation              & Input Size        & Output Size       & Kernel Size & Pad Size   & Pool Size  & Activation \\ \hline
1        & Convolution            & 1@1536$\times$512 & 32@768$\times$256 & 3$\times$3  & 1$\times$1 & 2$\times$2 & Leaky ReLU \\
2        & Convolution            & 32@768$\times$256 & 64@384$\times$128 & 3$\times$3  & 1$\times$1 & 2$\times$2 & Leaky ReLU \\
3        & Convolution            & 64@384$\times$128 & 64@192$\times$64  & 3$\times$3  & 1$\times$1 & 2$\times$2 & Leaky ReLU \\ \hline
4        & Convolution            & 64@192$\times$64  & 64@96$\times$32   & 3$\times$3  & 1$\times$1 & 2$\times$2 & Leaky ReLU \\
5        & Convolution            & 64@96$\times$32   & 64@48$\times$16   & 3$\times$3  & 1$\times$1 & 2$\times$2 & Leaky ReLU \\
6        & Convolution            & 64@48$\times$16   & 64@24$\times$8    & 3$\times$3  & 1$\times$1 & 2$\times$2 & Leaky ReLU \\
7        & Convolution            & 64@24$\times$8    & 128@12$\times$4   & 3$\times$3  & 1$\times$1 & 2$\times$2 & Leaky ReLU \\
8        & Convolution            & 128@24$\times$8   & 64@12$\times$4    & 1$\times$1  & -          & -          & Leaky ReLU \\
9        & Transposed convolution & 64@12$\times$4    & 64@24$\times$8    & 4$\times$4  & 1$\times$1 & -          & Leaky ReLU \\
10       & Transposed convolution & 64@24$\times$8    & 64@48$\times$16   & 4$\times$4  & 1$\times$1 & -          & Leaky ReLU \\
11       & Transposed convolution & 64@48$\times$16   & 64@92$\times$32   & 4$\times$4  & 1$\times$1 & -          & Leaky ReLU \\
12       & Transposed convolution & 64@92$\times$32   & 64@192$\times$64  & 4$\times$4  & 1$\times$1 & -          & Leaky ReLU \\ \hline
13       & Convolution            & 128@192$\times$64 & 1@192$\times$64   & 1$\times$1  & -          & -          & Sigmoid    \\ \hline
\end{tabular}
\end{table}


%
\section{Conclusion} \label{conclusion}
In this work, a methodology for designing DCNN models for railway maintenance tasks was introduced. Formally, bounding box regression is replaced with contour finding on thresholded saliency maps. To the best of our knowledge, this approach is a first. This concept takes advantage of the unique nature of railway track images when compared to generic object detection challenges. FasteNet performs fastener detections on images of 1600$\times$512 at a speed of 110 FPS on an Nvidia GTX 1080 with up to 99.1\% precision using a very lightweight network.

As it stands, there are several limitations to FasteNet - susceptibility to perspective and size warping being a major one, limiting its use case to non-airborne mounted camera systems. FasteNet may also be overfitting to the dataset, given that the dataset is considered small, it may not fully capture the distribution of such an application in real life.

In the future, several things will be looked at. Firstly, there is hope to make the dataset public and contain more samples. Second, detecting fasteners alone is not particularly useful. FasteNet will aim to be applicable to fastener, sleeper, rail, and ballast defects by having one extra channel per class of detection. The goal is leverage the principles of FasteNet to develop similar models for railway inspection and/or road inspection. FasteNet can also be adapted to perform multitask classification as in the case of \cite{gibert2016deep}. Additionally, an expansion of the dataset would allow the network to generalize better. Data augmentation methods such as a range of affine transforms, lux manipulation, self adversarial training or simply gathering more data can be used.

\bibliography{references}

\end{document}